\documentclass[11pt]{article}

\usepackage[preprint]{acl}

\usepackage{newtxtext}
\usepackage{latexsym}
\usepackage[T1]{fontenc}
\usepackage[utf8]{inputenc}
\usepackage{microtype}
\usepackage{inconsolata}
\usepackage{graphicx}
\usepackage{subcaption}
\usepackage{booktabs}
\usepackage{multirow}
\usepackage{pifont}
\usepackage{amsmath}
\usepackage{amssymb}
\usepackage{mathtools}
\usepackage{amsthm}
\usepackage[capitalize,noabbrev]{cleveref}

\theoremstyle{plain}

\theoremstyle{definition}

\theoremstyle{remark}

\title{Unified Visual-to-Code Generation via Symbolic Rewards and Reference-Guided Code Optimization}

\author{
  \textbf{Yaozhi Zheng\textsuperscript{1,*}},
  \textbf{Yilei Jiang\textsuperscript{1,*}},
  \textbf{Manyuan Zhang\textsuperscript{1}},
  \textbf{Yuxuan Wan\textsuperscript{1}},
\\
  \textbf{Kaituo Feng\textsuperscript{1}},
  \textbf{Tianshuo Peng\textsuperscript{1,2}},
  \textbf{Bo Zhang\textsuperscript{2}},
  \textbf{Xiangyu Yue\textsuperscript{1,\textdagger}}
\\
\\
  \textsuperscript{1}The Chinese University of Hong Kong
\\
  \textsuperscript{2}Shanghai Artificial Intelligence Laboratory
\\
  \small{
    \{zhengyz, yljiang\}@link.cuhk.edu.hk, xyyue@ie.cuhk.edu.hk
  }
}

\begin{document}
\maketitle
\begingroup
\renewcommand{\thefootnote}{\fnsymbol{footnote}}
\footnotetext[1]{Equal contribution.}
\footnotetext[2]{Corresponding author.}
\endgroup
\begin{abstract}
Visual-to-Code generation, which transforms scientific plots, vector graphics, and webpages into executable scripts, demands a level of pixel-precise alignment that standard Multimodal Large Language Models (MLLMs) fail to achieve through Supervised Fine-Tuning (SFT) alone. While Reinforcement Learning (RL) offers a theoretical pathway to bridge this gap, its application is hindered by two fundamental obstacles: (1) \textit{Reward Coarseness}, where semantic metrics like CLIP scores fail to penalize fine-grained element deviations, and (2) \textit{Exploration Stagnation}, where the sparse, heterogeneous code search space prevents the policy from bootstrapping valid trajectories. To overcome these limitations, we introduce UniCoder, a unified RL framework that integrates two novel mechanisms. First, we propose \textbf{Symbolic Attribute Alignment}, which employs a lightweight auxiliary LLM to parse generated code into discrete visual attributes (e.g., hex colors, coordinate limits), enabling dense, element-wise reward computation. Second, to escape local optima, we devise \textbf{Reference-Guided Code Optimization}, a strategy that dynamically injects ground-truth trajectories into low-performing rollout groups, transforming blind exploration into guided policy improvement. Extensive experiments on ChartMimic, UniSVG, Design2Code and ScreenBench benchmarks demonstrate that our 8B-parameter model not only surpasses all open-source baselines but also achieves state-of-the-art performance comparable to proprietary models, establishing a new paradigm for generalized visual-to-code synthesis. Our code will be open-sourced at \url{https://github.com/JimmyZhengyz/unicoder}.
\end{abstract}

\section{Introduction}
\label{sec:intro}

The capability to translate visual information into executable code stands as a pivotal frontier in Multimodal Large Language Model (MLLM) research. From reverse-engineering scientific charts into Python scripts to converting design mockups into functional webpages, this task bridges the gap between visual intent and digital implementation. Unlike general visual captioning where semantic ambiguity is frequently tolerated, visual-to-code generation demands rigorous precision. The generated code must not only be syntactically executable but also render a visual output that is structurally and stylistically isomorphic to the input.

Recent advancements in MLLMs have been largely driven by large-scale Supervised Fine-Tuning (SFT). While SFT enables models to learn high-level correspondences between image layouts and code structures, it often plateaus in fine-grained alignment. SFT-trained models frequently struggle with precise coordinate reasoning in Scalable Vector Graphics (SVG) or specific attribute extraction in scientific plotting. Consequently, they often default to hallucinated values that fit the training distribution rather than the specific instance.

Reinforcement Learning (RL) offers a promising paradigm to transcend these limitations by optimizing directly for outcome-based feedback. Techniques such as Group Relative Policy Optimization (GRPO) have demonstrated remarkable success in logical reasoning and code generation. However, adapting RL to the visual-to-code domain introduces two fundamental challenges: \textbf{(1) The Reward Alignment Problem:} Constructing a reward signal that is both dense and accurate is notoriously difficult. Pixel-level metrics are overly sensitive to high-frequency rendering artifacts where a minor padding shift can penalize a structurally correct solution. Conversely, semantic metrics are often too coarse to penalize hallucinations in text labels. \textbf{(2) Exploration Stagnation:} The search space for executable code is sparse and brittle. A model facing a complex visual input often suffers from a ``cold start'' problem where it fails to generate any executably valid trajectories during early training. This leads to gradient variance collapse and learning stagnation.

To address these challenges, we propose a unified reinforcement learning framework designed to master diverse visual-to-code tasks simultaneously. We introduce \textbf{Symbolic Attribute Alignment}, a mechanism that leverages a lightweight and auxiliary LLM to parse generated code and extract discrete visual attributes such as fill colors and axis limits. This enables the calculation of precise and element-level rewards that decouple structural correctness from rendering noise. Furthermore, we propose \textbf{Reference-Guided Code Optimization} to mitigate exploration stagnation. This strategy dynamically injects ground-truth reference code into the policy's rollout buffer during low-performance episodes. This effectively bootstraps the exploration process by transforming the optimization landscape from sparse blind search to guided imitation.

Our main contributions are summarized as follows:
\begin{itemize}
    \setlength{\itemsep}{0pt}
    \setlength{\parsep}{0pt}
    \setlength{\topsep}{0pt}
    \item We identify the critical bottlenecks of reward coarseness and exploration stagnation when applying RL to multimodal code generation.
    \item We propose UniCoder, a novel framework integrating an LLM-based Attribute Extractor for fine-grained reward modeling and a Reference Injection strategy for stable policy exploration.
    \item We demonstrate that our 8B-parameter model achieves state-of-the-art performance among open-source models on multiple diverse benchmarks including ChartMimic, UniSVG, Design2Code and ScreenBench. Our method significantly closes the gap with proprietary models like GPT-5.
\end{itemize}

\section{Related Work}

\subsection{Multimodal Large Language Models for Visual-to-Code Generation}
The emergence of Multimodal Large Language Models (MLLMs) has significantly advanced the field of joint visual-text reasoning. Early explorations such as LLaVA\cite{liu2023visualinstructiontuning} laid the groundwork for aligning visual and textual representations by introducing a simple linear projection layer to map CLIP visual features into the LLaMA word embedding space. Qwen-VL\cite{bai2023qwenvlversatilevisionlanguagemodel} and DeepSeek-VL\cite{lu2024deepseekvlrealworldvisionlanguageunderstanding} were fine-tuned to further enhance high-resolution processing and OCR capabilities essential for reading code-related visual elements. Contemporary foundation models like GPT-5.1 and Gemini 3 Pro have dramatically expanded the frontiers of multimodal understanding, excelling at tasks that require generating high-fidelity code from complex imagery. Parallel efforts pursue unified reasoning across modalities, with OneThinker~\cite{Feng_2026_CVPR} consolidating image and video understanding into a single model. Beyond general-purpose reasoning, increasing attention has been dedicated to specialized domains. \textbf{Python Plots}: Early works like DePlot\cite{liu2023deplotoneshotvisuallanguage} converted charts into textual tables for downstream reasoning. More recently, Plot2Code\cite{wu2024plot2codecomprehensivebenchmarkevaluating} and ChartMimic\cite{yang2025chartmimicevaluatinglmmscrossmodal} introduced benchmarks requiring models to generate executable Python code (e.g., Matplotlib) that strictly reproduces the visual appearance of reference charts. \textbf{SVG}: DeepSVG pioneered deep generative models for vector icons using hierarchical transformers to predict path commands. Building on this, IconShop\cite{wu2023iconshoptextguidedvectoricon} and StarVector\cite{rodriguez2025starvectorgeneratingscalablevector} explored text-guided SVG generation. For large-scale icon-to-SVG translation evaluation, UniSVG\cite{Li_2025} established a comprehensive framework highlighting the difficulty MLLMs face in maintaining topological correctness and geometric precision without explicit reinforcement. \textbf{Webpage}: WebSight\cite{bhathal2025websightvisionfirstarchitecturerobust} managed to utilize millions of synthetic screenshot-code pairs for fine-tuning. Design2Code\cite{si2025design2codebenchmarkingmultimodalcode} provided a rigorous benchmark for real-world frontend generation, demonstrating that while SFT models can capture high-level layouts, they often struggle with fine-grained alignment and complex DOM tree nesting. Recent attempts such as LayoutCoder integrated multi-agent framework to further enhance webpage fidelity for UI-to-Code generation. Beyond static frontend synthesis, agentic coding systems such as OpenGame~\cite{jiang2026opengameopenagenticcoding} extend executable code generation to interactive game environments.

\subsection{Reinforcement Learning for Multimodal Alignment}
Group Relative Policy Optimization (GRPO) \cite{shao2024deepseekmathpushinglimitsmathematical} has rapidly emerged as a paradigm-shifting approach for post-training alignment, demonstrating exceptional efficacy in unlocking complex reasoning capabilities, particularly for code generation tasks. Recent studies further show that group-based policy optimization can benefit from external gold references when sampled responses provide weak contrastive signals~\cite{wu2025gcpo}, suggesting the importance of reliable guidance in low-reward regimes. Complementary work explores dedicated reasoning reward models for agentic systems~\cite{fan2026exploringreasoningrewardmodel}, underscoring the broader challenge of designing reward functions that reflect task-specific correctness. However, applying such advanced optimization to domain-specific visual synthesis remains constrained by the quality of reward signals. Existing approaches in this field predominantly rely on coarse-grained objectives: Leading approaches for SVG generation like VectorFusion\cite{jain2022vectorfusiontexttosvgabstractingpixelbased} and StarVector\cite{rodriguez2025starvectorgeneratingscalablevector} predominantly rely on basic semantic distillation (e.g., CLIP Score) or coordinate reconstruction . ScreenCoder\cite{jiang2025screencoderadvancingvisualtocodegeneration} employed a pixel-level Mean Squared Error (MSE) as the primary reward signal during RL stage for Webpage-to-Code generation. These heuristic metrics, however, fail to capture element-level nuances due to the non-linear mapping between symbolic code and visual output, where subtle deviations in the code structure can trigger cascading layout failures.

\begin{figure*}[t]
    \centering
    \includegraphics[width=\linewidth]{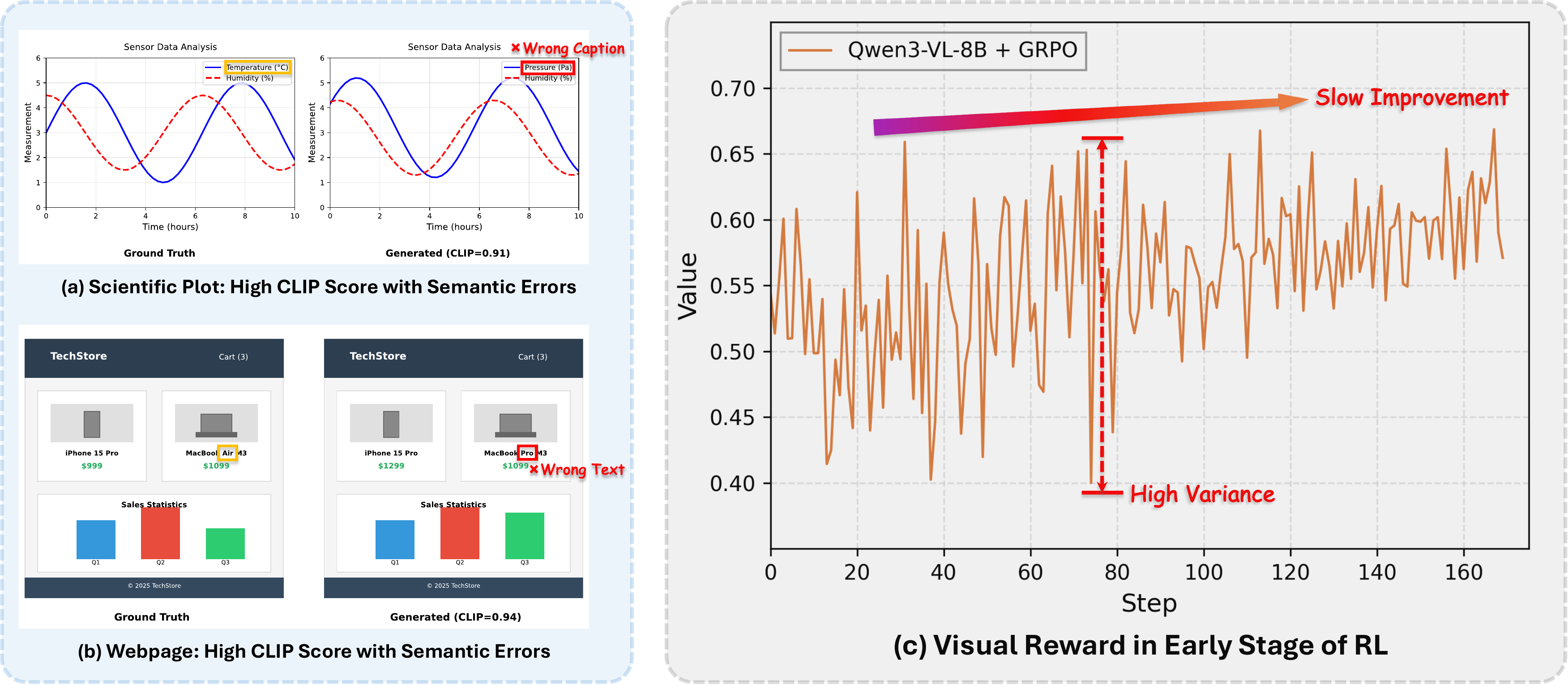}
    \caption{Limitations of CLIP-based visual rewards for Visual-to-Code generation. (a-b) \textbf{Semantic Error Cases}: Generated outputs receive high CLIP scores (0.91 and 0.94) despite containing critical errors, such as wrong axis labels and shifted data points in scientific plots, incorrect prices and product names in webpages. These examples demonstrate that CLIP captures global visual similarity but fails to enforce fine-grained element-level alignment. (c) \textbf{Exploration Stagnation}: Training with vanilla GRPO shows reward stagnation and high variance, indicating the model struggles to explore effectively across heterogeneous coding paradigms (Python/SVG/HTML).}
    \label{fig:motivation}
\end{figure*}

\section{Motivation}
\label{sec:motivation}

Developing a generalist agent for visual-to-code generation requires us to overcome two fundamental bottlenecks. These challenges stem from the unique, non-linear nature of translating visual intent into executable logic. We argue that standard reinforcement learning paradigms are ill-suited for this domain without specific structural adaptations.

\subsection{The Reward Alignment Paradox}
The first obstacle lies in the difficulty of defining a reward signal that is both chemically dense and semantically accurate. Unlike general text generation where semantic equivalence is broad, visual-to-code generation requires a rigorous isomorphism between the input image and the rendered output. We observe that conventional reward mechanisms fail to capture this relationship due to two distinct failure modes.

\paragraph{Semantic Coarseness of Visual Embeddings.} 
Metrics derived from pre-trained vision-language models, such as the CLIP Score, are optimized for global semantic alignment rather than fine-grained structural precision. As illustrated in Figure \ref{fig:motivation}(a) and (b), we observe that these metrics exhibit a dangerous tolerance for ``semantic hallucinations.'' A generated chart may receive a high similarity score because it successfully mimics the color scheme and general layout of the ground truth. However, it may simultaneously fail to render the correct data points or text labels. This creates a reward signal that encourages the model to cheat. The policy learns to prioritize aesthetic style over data fidelity.

\paragraph{Brittleness of Pixel-Level Objectives.} 
Conversely, attempting to enforce precision via pixel-level metrics like Mean Squared Error (MSE) introduces the opposite problem. The rendering process of code is highly non-linear and sensitive to high-frequency noise. A semantically irrelevant change, such as a minor shift in margin padding or a difference in font anti-aliasing, can result in a catastrophic pixel loss. This noisy feedback confuses the policy. It penalizes structurally correct code for rendering artifacts that are invisible to the human eye. This dichotomy motivates our design of \textbf{Symbolic Attribute Alignment}. We posit that the only way to accurately evaluate code is to decouple the \textit{symbolic intent} (what elements should exist) from the \textit{rendering realization} (how pixels are drawn).

\subsection{Exploration Stagnation in Heterogeneous Spaces}
The second major bottleneck is the difficulty of effective exploration. While Group Relative Policy Optimization (GRPO) has shown success in math and logic tasks, applying it to multimodal code generation reveals a severe ``cold start'' problem.

\paragraph{The Sparse Reward Landscape.} 
The search space for executable code is notoriously brittle. Unlike natural language where a slightly incorrect sentence is still readable, a slightly incorrect code block often results in a syntax error or a blank render. This creates an extremely sparse reward landscape. During the initial stages of training, the model frequently fails to generate any executable trajectories. Consequently, the gradient variance collapses. As shown in Figure \ref{fig:motivation}(c), this leads to prolonged periods of reward stagnation where the model learns nothing because it receives no positive feedback to reinforce.

\paragraph{Cognitive Burden of Heterogeneity.} 
This stagnation is exacerbated by the requirement to master multiple coding paradigms simultaneously. A unified model must navigate vastly different reasoning modalities. These include the logic-driven imperative programming of Python, the geometry-driven coordinate systems of SVG, and the structure-driven DOM nesting of HTML. We observe that standard random exploration struggles to toggle between these modalities effectively. The model often gets stuck in local optima where it outputs valid code for the wrong domain or hallucinates syntax from one language into another. This observation serves as the foundation for our \textbf{Reference-Guided Optimization}. We argue that to traverse such a disjoint and punishing search space, the optimization process must be transformed from a blind search into a guided discovery.

\begin{figure*}[t]
    \centering
    \includegraphics[width=\textwidth]{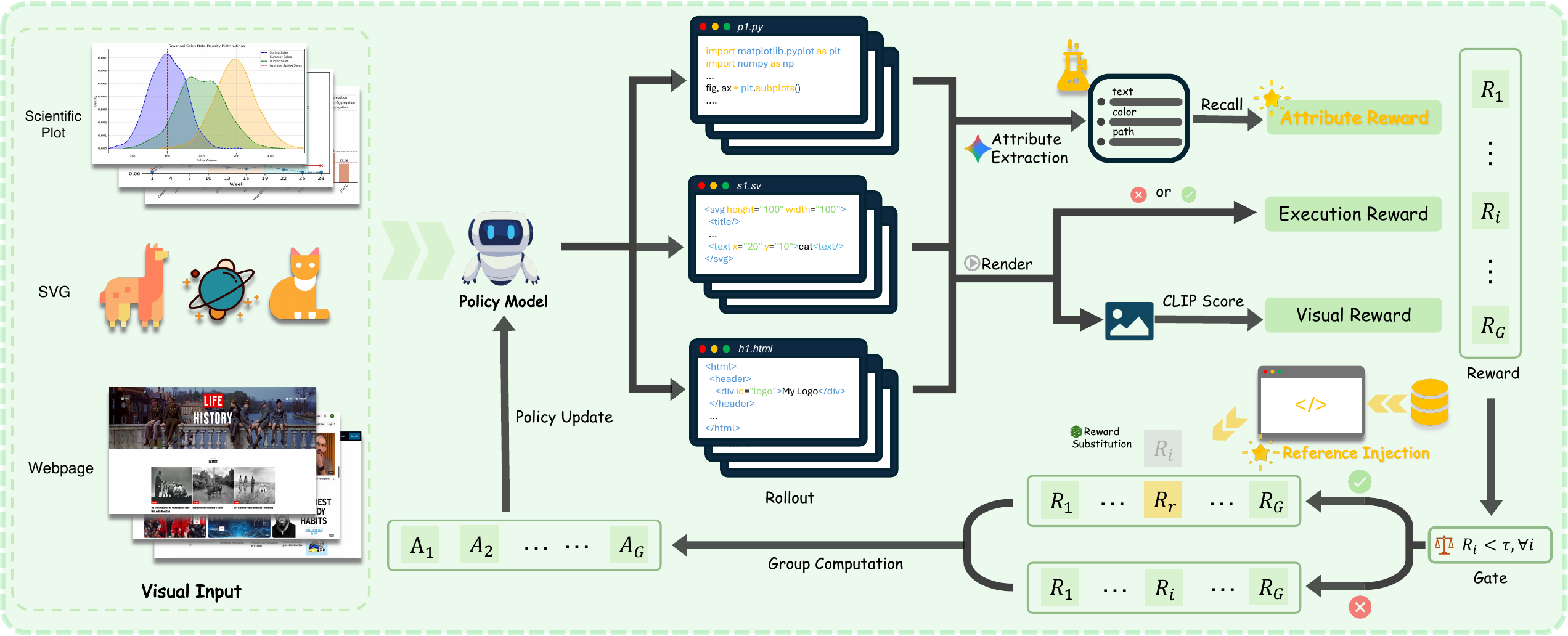} 
    \caption{\textbf{Overview of the proposed framework.} The model processes diverse visual inputs (Scientific Plots, SVGs, Webpages) to generate corresponding executable code. To ensure high-quality generation, we employ a multi-faceted reward mechanism consisting of: (1) \textbf{Attribute Reward} for fine-grained symbolic alignment, (2) \textbf{Execution Reward} to verify code compilability, and (3) \textbf{Visual Reward} (e.g., CLIP score) for semantic consistency. These rewards guide the \textit{Policy Update} via a reference-guided group computation strategy.}
    \label{fig:main_framework}
\end{figure*}

\section{Method}
\label{sec:method}

We propose a unified reinforcement learning framework designed to overcome the reward sparsity and exploration stagnation inherent to multimodal code generation. Our approach, Reference-Guided Group Relative Policy Optimization (RG-GRPO), adapts the standard GRPO paradigm by introducing two critical modifications: (1) a \textit{Symbolic Attribute Alignment} mechanism that densifies the reward signal, and (2) a \textit{Reference Injection} strategy that dynamically alters the exploration distribution to bootstrap learning. The overall framework is illustrated in Figure \ref{fig:main_framework}.

\subsection{Problem Formulation}
We consider the task of visual-to-code generation as a conditional generation problem. Given a visual input $x$ (e.g., a chart image, SVG, or webpage screenshot) and a textual instruction $I$, the goal is to generate executable code $y = (y_1, ..., y_T)$ that renders into an image $\mathcal{R}(y)$ visually isomorphic to $x$.
Standard Supervised Fine-Tuning (SFT) optimizes the likelihood $P_\theta(y|x, I)$. However, SFT lacks the feedback mechanism to penalize visually minor but symbolically fatal errors. We therefore transition to Reinforcement Learning, aiming to maximize an expected reward:
\begin{equation}
    J(\theta) = \mathbb{E}_{y \sim \pi_\theta(\cdot|x, I)} [R(y, x)]
\end{equation}
where $R$ is our proposed multi-faceted reward function.

\subsection{Multi-Faceted Reward Modeling}
\label{subsec:reward}
To address the ``Reward Alignment Paradox'' (Section \ref{sec:motivation}), we construct a composite reward function that decouples executability, semantic fidelity, and symbolic precision.

\paragraph{Visual and Execution Feedback.}
We employ two foundational signals:
\begin{itemize}
    \setlength{\itemsep}{0pt}
    \setlength{\parsep}{0pt}
    \setlength{\topsep}{0pt}
    \item \textbf{Execution Constraints ($R_{exec}$):} A binary indicator where $R_{exec}(y) = 1$ if $\mathcal{R}(y) \neq \emptyset$ (i.e., the code compiles and renders), and 0 otherwise. This acts as a hard gatekeeper for the reward signal.
    \item \textbf{Semantic Similarity ($R_{vis}$):} For valid executions, we compute the cosine similarity between the CLIP embeddings of the rendered image $\mathcal{R}(y)$ and the ground truth $x$: $R_{vis} = \text{CLIP}(\mathcal{R}(y), x)$.
\end{itemize}

\paragraph{Symbolic Attribute Alignment.}
Standard visual metrics fail to capture fine-grained details (e.g., exact hex colors or axis ranges). To bridge this gap, we introduce an \textbf{Attribute Extraction Module}, $\mathcal{E}_\phi$, parameterized by a lightweight auxiliary LLM (e.g., Qwen-2.5-3B).
For a generated code sequence $y$, the module extracts a discrete set of symbolic attributes $A_{gen} = \mathcal{E}_\phi(y)$. This set includes domain-specific elements: $\{ \text{color\_codes}, \text{text\_labels}, \text{axis\_limits} \}$ for plots, or $\{ \text{DOM\_tags}, \text{styles} \}$ for webpages.

We define the Attribute Reward $R_{attr}$ as the \textit{Soft Recall} of these generated attributes against the ground-truth attributes $A_{gt} = \mathcal{E}_\phi(y_{gt})$:
\begin{equation}
    R_{attr}(y, y_{gt}) = \frac{\sum_{a \in A_{gen}} \mathbb{I}(a \in A_{gt})}{|A_{gt}| + \epsilon}
\end{equation}
Unlike exact matching, this formulation allows partial credit (e.g., getting the colors right but the labels wrong), providing a dense gradient signal even when the global structure is imperfect.

The final reward is a weighted sum: $R_{total} = w_1 R_{exec} + w_2 R_{vis} + w_3 R_{attr}$.

\subsection{Reference-Guided Exploration Strategy}
A standard GRPO update samples a group of outputs $G = \{y_1, ..., y_G\} \sim \pi_\theta(x)$ and computes advantages relative to the group mean. However, in the visual-to-code domain, the "cold start" problem often results in a group where \textit{all} candidates fail ($R_{total} \approx 0$). In this scenario, the group mean is zero, the variance is zero, and the gradient becomes uninformative.

To mitigate this, we propose a Reference Injection strategy that modifies the sampling distribution. During the rollout phase, we monitor the quality of the generated group. If the maximum reward in the group falls below a threshold $\tau$, we force-replace the lowest-performing candidate $y_{worst}$ with the ground-truth reference code $y_{ref}$.
\begin{equation}
G' = 
\begin{cases} 
(G \setminus \{y_w\}) \cup \{y_{\text{ref}}\}, \\
\quad \text{if } \max_{y \in G} R(y) < \tau, \\[1ex]
G, \\
\quad \text{otherwise}.
\end{cases}
\end{equation}
This injection transforms the learning process from pure exploration to \textit{dynamic imitation}. The reference code $y_{ref}$ introduces a high-reward trajectory into the group, ensuring that the group mean $\bar{R}$ is non-trivial.

\subsection{Reference-Aware Policy Optimization}
Our optimization objective is a modification of the GRPO loss. We utilize the injected group $G'$ to compute the advantage estimates. For each sample $y_i \in G'$, the advantage is standardized as:
\begin{equation}
    A_i = \frac{R(y_i) - \frac{1}{|G'|} \sum_{y_j \in G'} R(y_j)}{\text{std}(\{R(y_j)\}_{j=1}^{|G'|}) + \epsilon}
\end{equation}

\paragraph{The Anchor Effect.}
The inclusion of $y_{ref}$ (where $R(y_{ref}) \gg 0$) critically alters the optimization dynamics. It acts as a \textbf{performance anchor}, significantly raising the group mean. Consequently, sub-optimal generated samples $y_{gen}$ receive a strongly negative advantage $A_{gen} \ll 0$.
The objective function is then:
\begin{equation}
\begin{aligned}
    \mathcal{L}_{\text{RG-GRPO}} &= - \frac{1}{|G'|} \sum_{y_i \in G'} \Bigg[ \\
    &\quad \min \bigg( \frac{\pi_\theta(y_i)}{\pi_{\text{old}}(y_i)} A_i, \text{clip}(\dots) A_i \bigg) \\
    &\quad - \beta \mathbb{D}_{KL} \Bigg]
\end{aligned}
\end{equation}
where $\rho_i = \pi_\theta(y_i) / \pi_{\mathrm{old}}(y_i)$. By ``anchoring'' the advantage calculation with the reference, the optimizer is not only encouraged to increase the probability of the injected reference (imitation) but is also explicitly discouraged from repeating the failed exploration attempts (suppression of errors), leading to faster convergence and reduced variance compared to vanilla GRPO.

\begin{table*}[t]
  \begin{center}
      \resizebox{\textwidth}{!}{
      \begin{tabular}{l|ccc|ccc|ccccc|ccccc}
        \toprule
        \multirow{2}{*}{Model} & \multicolumn{3}{c|}{ChartMimic\_direct\_v2} & \multicolumn{3}{c|}{UniSVG-ISVGEN} & \multicolumn{5}{c|}{Design2Code} & \multicolumn{5}{c}{ScreenBench} \\
        \cmidrule(lr){2-4} \cmidrule(lr){5-7} \cmidrule(lr){8-12} \cmidrule(lr){13-17}
         & Exec.R & Low-L & High-L & Low-L & High-L & Score & Block & Text & Pos. & Color & CLIP & Block & Text & Pos. & Color & CLIP \\
        \midrule
        \multicolumn{17}{l}{Closed-Source Models} \\
        \midrule
        Gemini-3-Pro       & 83.2 & 71.5 & 70.8 & \underline{71.5} & 84.8 & 80.4 & 0.852 & 0.971 & 0.905 & 0.892 & 0.918 & 0.752 & 0.851 & 0.735 & 0.712 & 0.798 \\
        Claude-4.5-Opus   & \underline{85.8} & 73.2 & 72.5 & 69.5 & 85.5 & 81.2 & \underline{0.858} & 0.969 & \underline{0.912} & 0.885 & 0.915 & \underline{0.761} & 0.849 & \underline{0.742} & \underline{0.725} & \underline{0.805} \\
        GPT-5             & 84.5 & \underline{75.6} & \underline{73.9} & 70.8 & \underline{88.8} & \underline{81.9} & 0.855 & \underline{0.973} & 0.898 & \underline{0.899} & \underline{0.920} & 0.758 & \underline{0.854} & 0.728 & 0.718 & 0.802 \\
        \midrule
        \multicolumn{17}{l}{Open-Source Models} \\
        \midrule
        LLaVA-v1.6-7B   & 60.2 & 20.4 & 20.7 & 43.2 & 74.6 & 62.0 & 0.736 & 0.910 & 0.729 & 0.816 & 0.802 & 0.635 & 0.830 & 0.544 & 0.592 & 0.727 \\
        InternVL3.5-8B  & 66.5 & 42.4 & 47.1 & 54.1 & 76.9 & 67.8 & 0.805 & 0.942 & 0.798 & 0.825 & 0.876 & 0.710 & 0.822 & 0.598 & 0.620 & 0.755 \\
        Deepseek-VL-7B  & 41.1 & 18.8 & 20.5 & 47.2 & 73.5 & 63.0 & 0.718 & 0.824 & 0.702 & 0.720 & 0.843 & 0.680 & 0.773 & 0.570 & 0.614 & 0.732 \\
        Qwen2.5-VL-7B   & 65.5 & 39.9 & 40.7 & 56.4 & 73.8 & 63.3 & 0.822 & 0.951 & 0.815 & 0.831 & 0.893 & 0.723 & 0.828 & 0.613 & 0.632 & 0.762 \\
        Qwen3-VL-8B     & 77.3 & 63.4 & 66.8 & 60.0 & 78.1 & 70.7 & 0.841 & 0.968 & 0.852 & 0.855 & 0.905 & 0.741 & 0.843 & 0.652 & 0.661 & 0.784 \\
        Ours      & \textbf{86.4} & \textbf{76.1} & \textbf{74.4} & \textbf{72.2} & \textbf{89.1} & \textbf{82.3} & \textbf{0.865} & \textbf{0.975} & \textbf{0.925} & \textbf{0.908} & \textbf{0.922} & \textbf{0.768} & \textbf{0.857} & \textbf{0.755} & \textbf{0.734} & \textbf{0.812} \\
        \bottomrule
      \end{tabular}
      }
  \end{center}
  \caption{Comparison of different models on ChartMimic, UniSVG, Design2Code, and ScreenBench benchmarks. Best results are highlighted in \textbf{bold} and second best are \underline{underlined}.}
  \label{tab:model-comparison-mini}
  \vskip -0.1in
\end{table*}

\begin{figure*}[t]
    \centering
    \includegraphics[width=0.85\textwidth]{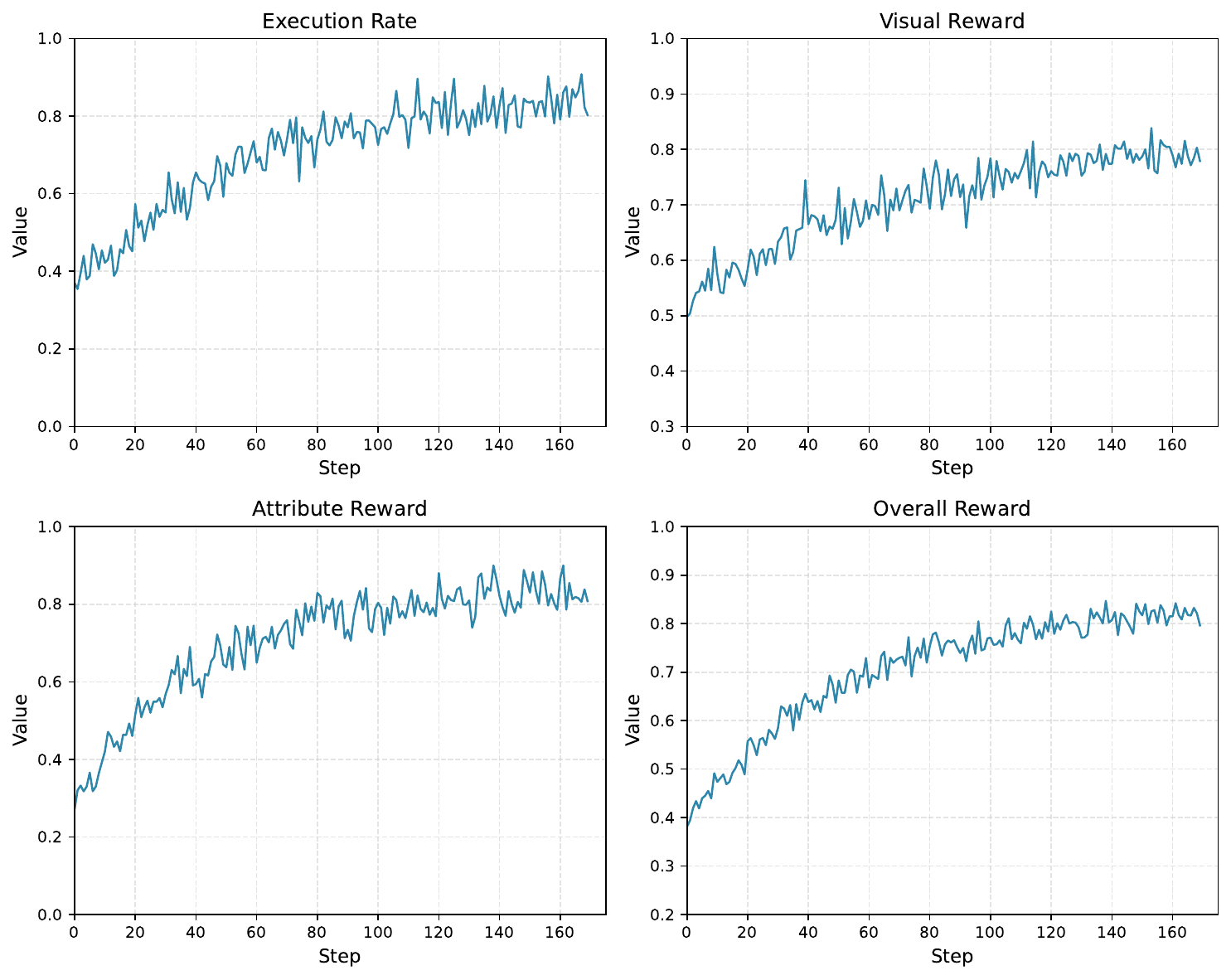}
    \caption{Training reward curves of our method. We report four metrics: (a) \textbf{Execution Rate} measures the fraction of generated code that compiles and renders successfully; (b) \textbf{Visual Reward} captures the CLIP similarity between rendered outputs and ground truth images; (c) \textbf{Attribute Reward} quantifies the element-level alignment via our proposed attribute extraction mechanism; (d) \textbf{Overall Reward} is the weighted combination of the three components ($R_{total} = w_1 R_{exec} + w_2 R_{vis} + w_3 R_{attr}$). All metrics show consistent improvement throughout training, demonstrating the effectiveness of our Reference-Guided GRPO framework.}
    \label{fig:training_rewards}
\end{figure*}

\section{Experiments}
\subsection{Experimental Setup}
\label{subsec:setup}
\paragraph{Datasets and Benchmarks.}
We conduct comprehensive experiments to evaluate the unified multimodal code generation capabilities of our framework across three distinct domains: scientific Python plotting, scalable vector graphics (SVG) creation, and webpage code generation.
\begin{itemize}
    \setlength{\itemsep}{0pt}
    \setlength{\parsep}{0pt}
    \setlength{\topsep}{0pt}
    \item \textbf{Training Data:} To foster robust cross-task generalization, we curate a unified instruction tuning dataset comprising \textbf{MSRL}~\cite{chen2025breakingsftplateaumultimodal} for Python plot generation, \textbf{SVG-Stack}~\cite{rodriguez2025starvectorgeneratingscalablevector} for SVG creation, and \textbf{Web2Code}~\cite{yun2024web2codelargescalewebpagetocodedataset} for hierarchical webpage structures.
    \item \textbf{Evaluation Benchmarks:} We assess performance on authoritative domain-specific benchmarks: \textbf{ChartMimic}~\cite{yang2025chartmimicevaluatinglmmscrossmodal} for chart-to-code generation (under the direct generation setting), \textbf{UniSVG}~\cite{Li_2025} for SVG translation, and \textbf{Design2Code}~\cite{si2025design2codebenchmarkingmultimodalcode} and \textbf{Screen-10K}~\cite{jiang2025screencoderadvancingvisualtocodegeneration}for real-world webpage-to-HTML synthesis. For ChartMimic and UniSVG, we report the official low-level and high-level metrics used by each benchmark. For Design2Code and ScreenBench, we report block, text, position, color, and CLIP-based similarity metrics.
\end{itemize}

\paragraph{Implementation Details.}
\textbf{Model Architecture:} We adopt \texttt{Qwen3VL-8B-Instruct} as our backbone policy model, selected for its strong multimodal reasoning capabilities among open-source MLLMs. For the Reference-Guided reward calculation, we utilize the lightweight \texttt{Qwen2.5-3b-Instruct} specifically as the Attribute Extractor. This smaller model is selected to strike an optimal balance between computational efficiency and the structural understanding required for feature extraction.

\noindent \textbf{Training Configuration:} Our Visual Reinforcement Learning stage is built upon the Group Relative Policy Optimization (GRPO) algorithm. We set the global batch size to 128 and the number of rollouts per prompt to $G=16$ to estimate the group baseline. During inference and rollout generation, we use a temperature of $0.7$ with a maximum response length of 4096 tokens.

\subsection{Main Results}
We evaluate our proposed method against state-of-the-art open-source MLLMs and proprietary closed-source models, as shown in Table~\ref{tab:model-comparison-mini}.

\paragraph{Superiority over Open-Source Baselines.} As shown in the table, our model, Qwen3-VL-8B (Ours), achieves state-of-the-art performance among all open-source models of similar scale. Compared to the strong baseline Qwen3-VL-8B, our method delivers substantial and consistent improvements across all metrics. This validates the effectiveness of our reinforcement learning framework in enhancing the visual-to-code generation capabilities of MLLMs. Notably, our model outperforms competing other 7B/8B class models (e.g., LLaVA-v1.6, InternVL3.5) by a significant margin, establishing a new benchmark for open-source performance in this domain.

\begin{table*}[t]
  \centering
  \resizebox{\textwidth}{!}{
  \begin{tabular}{l|ccc|ccc|ccccc|ccccc}
    \toprule
    \multirow{2}{*}{Configuration} & \multicolumn{3}{c|}{ChartMimic} & \multicolumn{3}{c|}{UniSVG} & \multicolumn{5}{c|}{Design2Code} & \multicolumn{5}{c}{ScreenBench} \\
    \cmidrule(lr){2-4} \cmidrule(lr){5-7} \cmidrule(lr){8-12} \cmidrule(lr){13-17}
     & Exec.R & Low-L & High-L & Low-L & High-L & Score & Block & Text & Pos. & Color & CLIP & Block & Text & Pos. & Color & CLIP \\
    \midrule
    Qwen3-VL-8B (Base) & 77.3 & 63.4 & 66.8 & 60.0 & 78.1 & 70.7 & 0.841 & 0.968 & 0.852 & 0.855 & 0.905 & 0.741 & 0.843 & 0.652 & 0.661 & 0.784 \\
    \midrule
    \multicolumn{17}{l}{\textit{Ablation Variants}} \\
    \ \ w/o Reference Injection & 79.6 & 69.8 & 68.5 & 68.8 & 85.6 & 78.9 & 0.848 & 0.970 & 0.895 & 0.885 & 0.910 & 0.748 & 0.846 & 0.710 & 0.705 & 0.792 \\
    \ \ w/o Attribute Reward & 83.2 & 70.4 & 71.3 & 67.5 & 81.7 & 76.2 & 0.859 & 0.972 & 0.875 & 0.870 & 0.915 & 0.762 & 0.852 & 0.685 & 0.680 & 0.801 \\
    \midrule
    \textbf{Ours (Full Model)} & \textbf{86.4} & \textbf{76.1} & \textbf{74.4} & \textbf{72.2} & \textbf{89.1} & \textbf{82.3} & \textbf{0.865} & \textbf{0.975} & \textbf{0.925} & \textbf{0.908} & \textbf{0.922} & \textbf{0.768} & \textbf{0.857} & \textbf{0.755} & \textbf{0.734} & \textbf{0.812} \\
    \bottomrule
  \end{tabular}
  }
  \vspace{4pt}
  \caption{Ablation study on ChartMimic, UniSVG, Design2Code, and ScreenBench. The results demonstrate that the Attribute Reward is particularly critical for fine-grained metrics (Position, Color), while Reference Injection significantly stabilizes layout performance (Block).}
  \label{tab:ablation-full}
\end{table*}

\paragraph{Comparison with Closed-Source Models.}
We further extend our evaluation to leading proprietary models, including Gemini 3 Pro, GPT 5 and Claude 4.5 Opus. Despite the substantial disparity in parameter scale and training resources, \texttt{Qwen3-VL-8B (Ours)} demonstrates remarkably competitive performance. As detailed in Table 1, our method significantly narrows the gap with state-of-the-art commercial systems on the Design2Code benchmark and even surpasses larger proprietary models on specific visual alignment tasks. This result indicates that our visual reinforcement learning framework effectively unlocks the potential of compact models, enabling them to rival large-scale systems in specialized multimodal code generation tasks without requiring massive computational overhead.

\paragraph{Training Analysis.}
To understand the optimization dynamics of our framework, we visualize the evolution of the reward components during training in Figure \ref{fig:training_rewards}.
\begin{itemize}
    \setlength{\itemsep}{0pt}
    \setlength{\parsep}{0pt}
    \setlength{\topsep}{0pt}
    \item \textbf{Syntactic Correctness:} The \textit{Execution Rate} (Figure \ref{fig:training_rewards}a) increases rapidly in the early stages, indicating that the policy quickly learns to generate syntactically valid code that renders successfully.
    \item \textbf{Visual \& Structural Alignment:} Both \textit{Visual Reward} and \textit{Attribute Reward} (Figure \ref{fig:training_rewards}b, c) show consistent growth, demonstrating that the model progressively learns to align the generated output with the visual constraints of the reference image.
    \item \textbf{Convergence:} The \textit{Overall Reward} (Figure \ref{fig:training_rewards}d) exhibits a stable upward trend, confirming that our Reference-Guided GRPO algorithm effectively optimizes the multi-objective objective without suffering from reward collapse or instability.
\end{itemize}

\subsection{Ablation Study}
\label{subsec:ablation}

To verify the effectiveness of each component in our proposed framework, we conduct ablation studies on the \texttt{Qwen3-VL-8B} backbone. We dissect the contributions of our two core innovations: the \textit{Reference Injection Strategy} and the \textit{Element-level Attribute Reward}.

\paragraph{Experimental Configurations.} We evaluate four distinct settings as shown in Table~\ref{tab:ablation-full}. First, the Base Model serves as the vanilla Qwen3-VL-8B baseline without our RL framework. Second, w/o Reference Injection tests our method excluding the injection mechanism (relying on standard prompting). Third, w/o Attribute Reward uses only the global CLIP score as the reward signal, removing the fine-grained attribute extractor. Finally, Ours (Full Model) represents the complete framework combining Reference Injection with the Attribute Reward.

\paragraph{Impact of Attribute Reward (Fine-Grained Alignment).} The ablation of the Attribute Reward reveals its specific role in refining local visual details. Comparing Ours with w/o Attribute Reward, we observe that while global metrics like CLIP scores remain relatively high, fine-grained structural metrics suffer significantly without element-level supervision. On Design2Code, the Position score drops from 0.925 to 0.875, and Color accuracy falls from 0.908 to 0.870. Similarly, on ScreenBench, removing the attribute reward causes a sharp decline in Position ($0.755 \to 0.685$) and Color ($0.734 \to 0.680$). These results confirm that while a global reward guides general semantic alignment, our proposed Attribute Extractor is essential for enforcing precise spatial layout and color fidelity.

\paragraph{Impact of Reference Injection (Stability).} The \textit{Reference Injection} strategy proves critical for stabilizing the generation process and ensuring layout correctness. Regarding execution stability on ChartMimic, removing the injection mechanism (w/o Reference Injection) leads to a significant drop in Execution Rate from 86.4\% to 79.6\%, indicating that the model struggles to generate valid plotting code without the structural hints provided by the injection. Furthermore, regarding layout consistency on Design2Code, the Block metric (measuring layout bounding box alignment) decreases from 0.865 to 0.848 without injection. This suggests that the Reference Injection strategy effectively constrains the infinite action space, guiding the model toward structurally valid and visually consistent outputs.




\subsection{Qualitative Result}
Additional qualitative comparisons are provided in Appendix~\ref{sec:additional_qualitative}, illustrating typical improvements in color fidelity, layout structure, and element completeness.

\section{Conclusion}
We presented UniCoder, a unified reinforcement learning framework for visual-to-code generation. UniCoder addresses reward coarseness and exploration stagnation through Symbolic Attribute Alignment and Reference-Guided Code Optimization, which provide dense element-level feedback and stabilize policy exploration with reference trajectories. Experiments across scientific plotting, SVG generation, and webpage synthesis show that our 8B model substantially improves over open-source baselines and narrows the gap with leading proprietary systems, indicating that fine-grained symbolic rewards and reference-guided exploration are effective for improving precise visual-to-code synthesis.


\section*{Limitations}

First, our evaluation focuses on three representative visual-to-code settings: scientific plotting, SVG generation, and webpage synthesis. These settings cover diverse rendering targets and programming structures, but they are still centered on static visual reconstruction. The current study does not address interactive interfaces, animations, mobile applications, or programs whose correctness depends on complex runtime behavior beyond the rendered static output. Second, our symbolic reward design relies on task-relevant visual attributes that can be extracted and compared from code, such as colors, text labels, coordinate ranges, and layout-related styles. While these attributes are well aligned with the benchmarks studied in this paper, extending the same framework to substantially different visual domains may require adapting the attribute schema and extraction prompts to capture new domain-specific properties.


\bibliography{main}

\clearpage
\appendix
\section{Additional Qualitative Comparisons}
\label{sec:additional_qualitative}

\begin{figure*}
    \centering
    \includegraphics[width=0.97\textwidth]{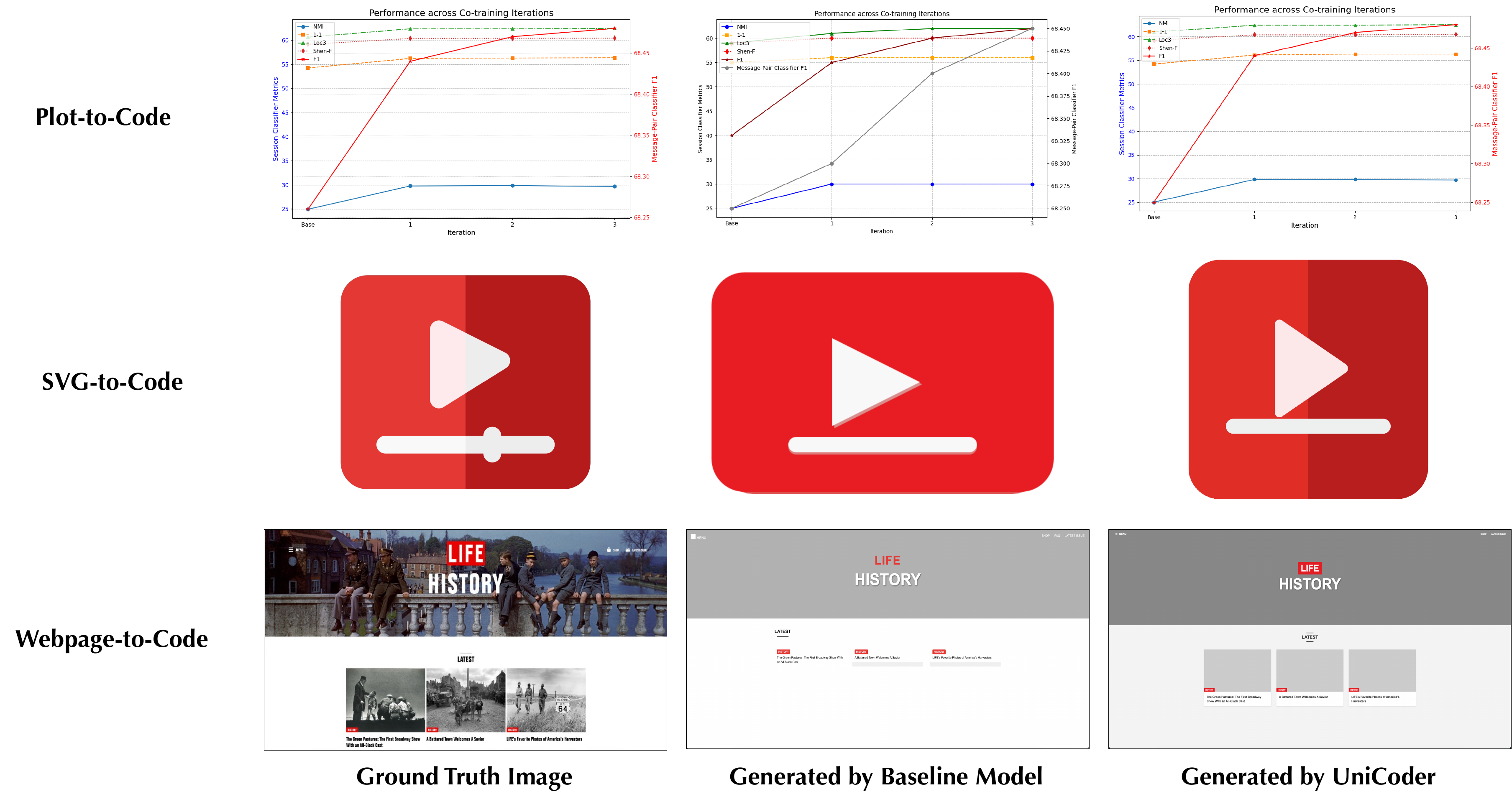}
    \caption{Qualitative comparison. As shown, the baseline model suffers from severe color hallucination and missing of elements, failing to replicate the visual palette. In contrast, our UniCoder achieves the highest fidelity, accurately recovering both the specific hex codes and the complete structure with only negligible errors.}
    \label{fig:qualitative results}
\end{figure*}

Figure \ref{fig:qualitative results} provides a qualitative comparison that highlights the practical advantages of our method. Standard baselines and even leading proprietary MLLMs demonstrate distinct \textit{perception} and \textit{planning} failures when handling complex plotting tasks. The baseline model suffers from severe perception alignment issues, failing to recognize specific visual attributes and reverting to default library color palettes. In contrast, UniCoder produces a high-fidelity result that closely mirrors both the stylistic appearance and the structural integrity of the reference, with only negligible deviation.

To further demonstrate the generalization capability of our framework, this appendix provides an extended set of qualitative comparisons. The following figures showcase side-by-side comparisons between the executable code outputs generated by our method, UniCoder, and those produced by leading baseline MLLMs such as Qwen3-VL, GPT-5, and Gemini-3-Pro. 

\begin{figure*}[t]
    \centering
    \includegraphics[width=0.85\textwidth]{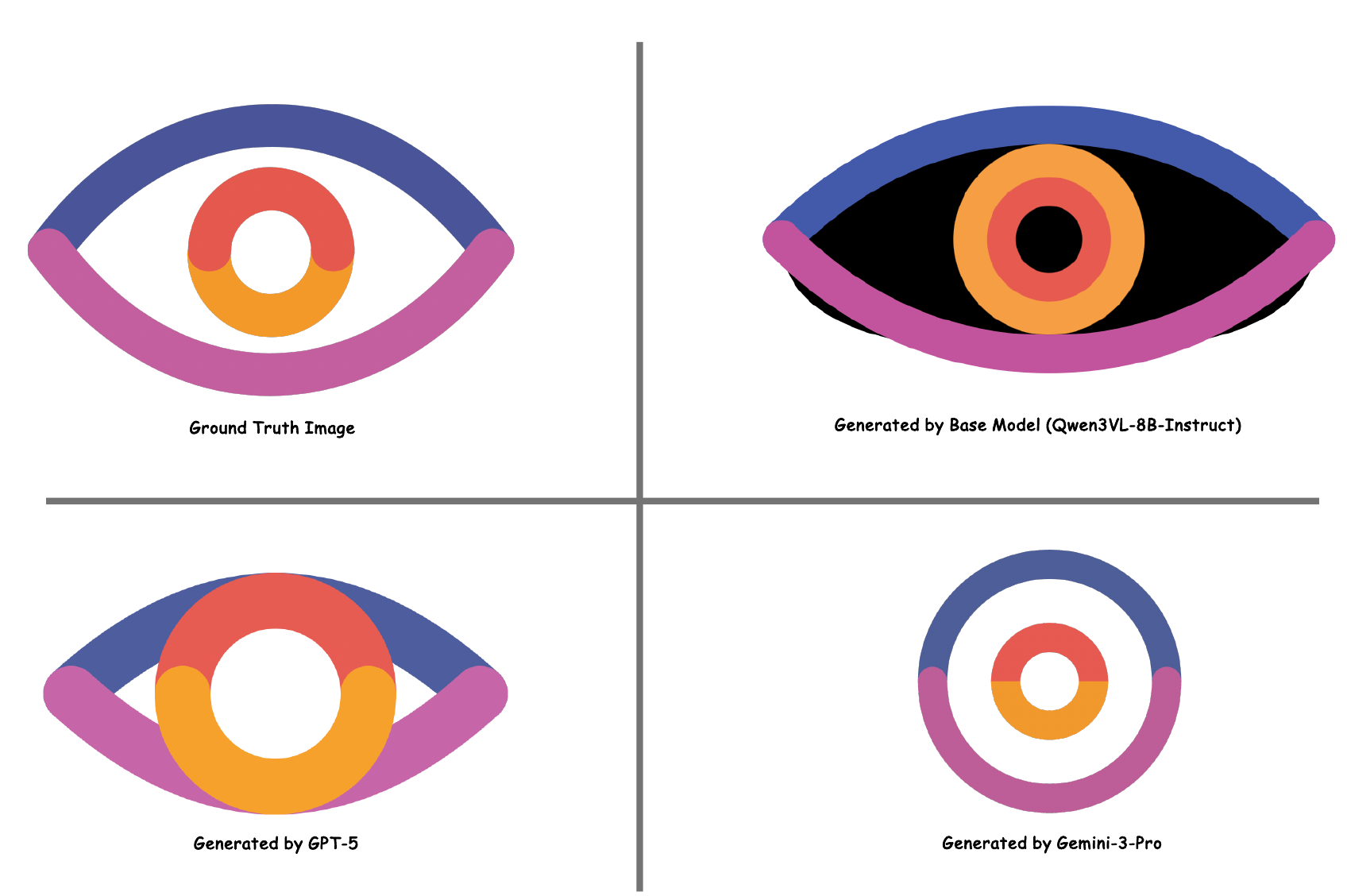}
    \caption{Qualitative comparison between our proposed method and various baselines.}
    \label{fig:case_2_r}
\end{figure*}

\begin{figure*}[t!]
    \centering
    \includegraphics[width=0.65\textwidth]{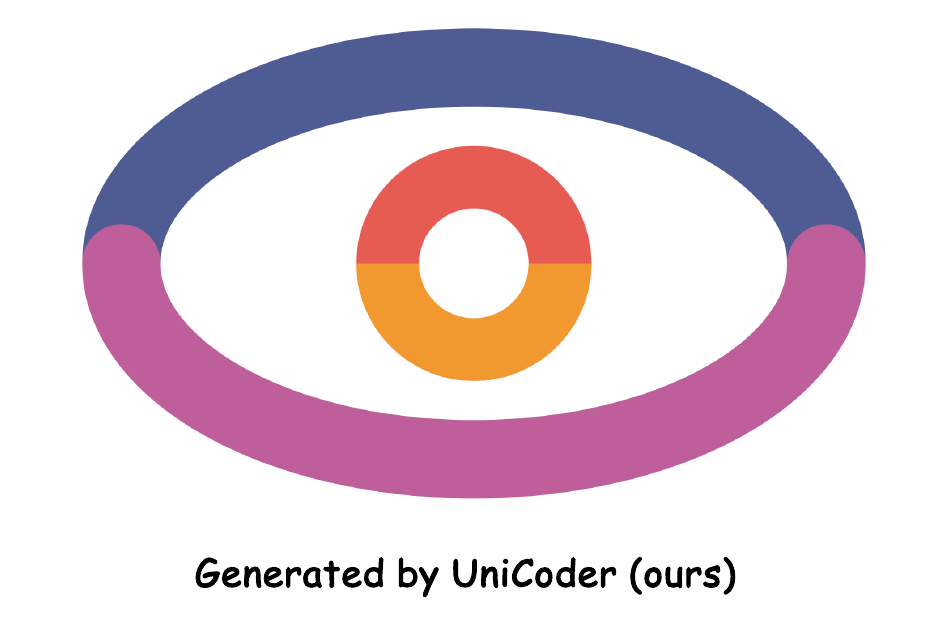}
    \caption{Qualitative comparison between our proposed method and various baselines.}
    \label{fig:case_2_ours}
\end{figure*}

\begin{figure*}[t]
    \centering
    \includegraphics[width=0.85\textwidth]{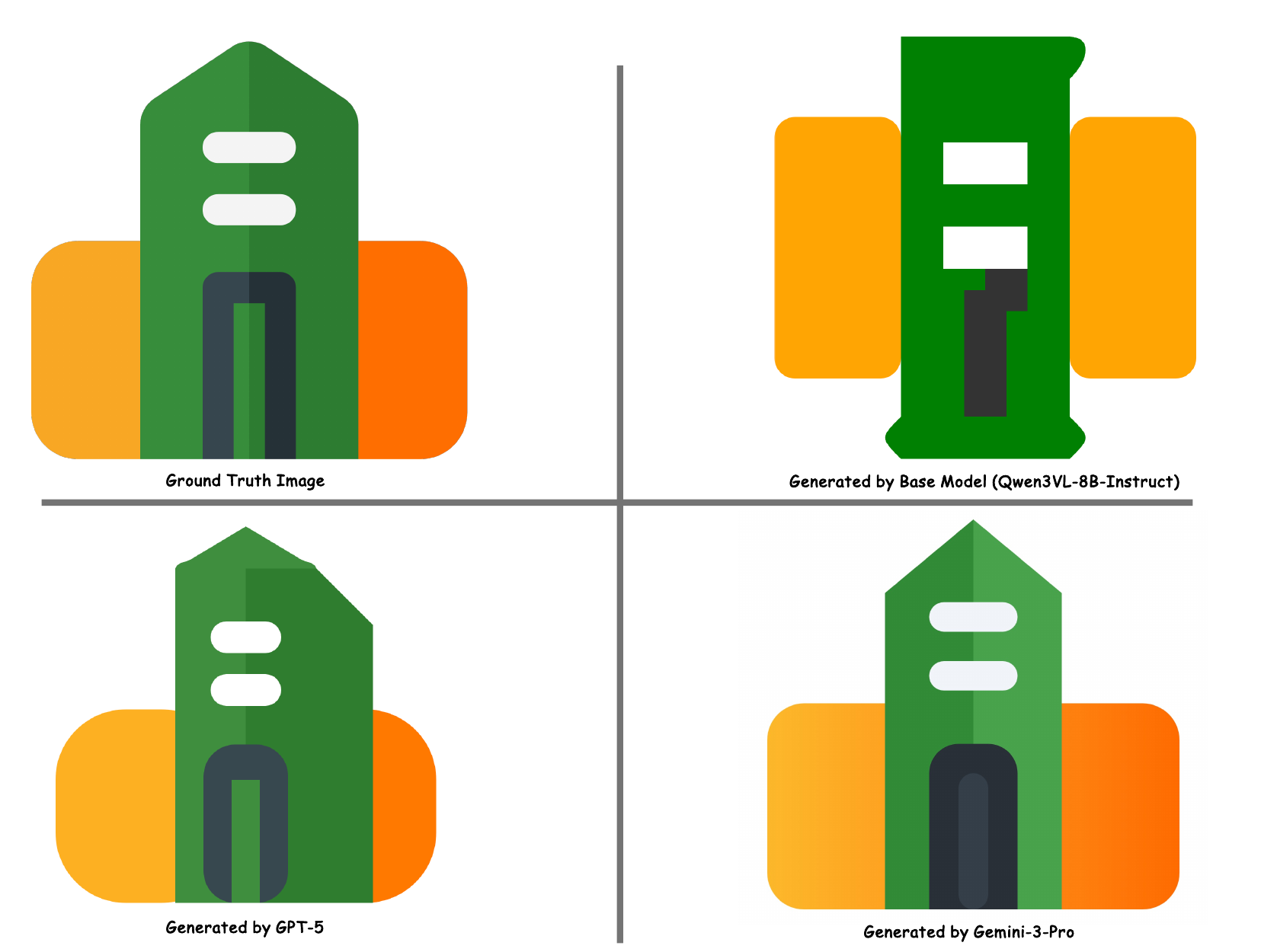}
    \caption{Qualitative comparison between our proposed method and various baselines.}
    \label{fig:case_3_r}
\end{figure*}

\begin{figure*}[t!]
    \centering
    \includegraphics[width=0.65\textwidth]{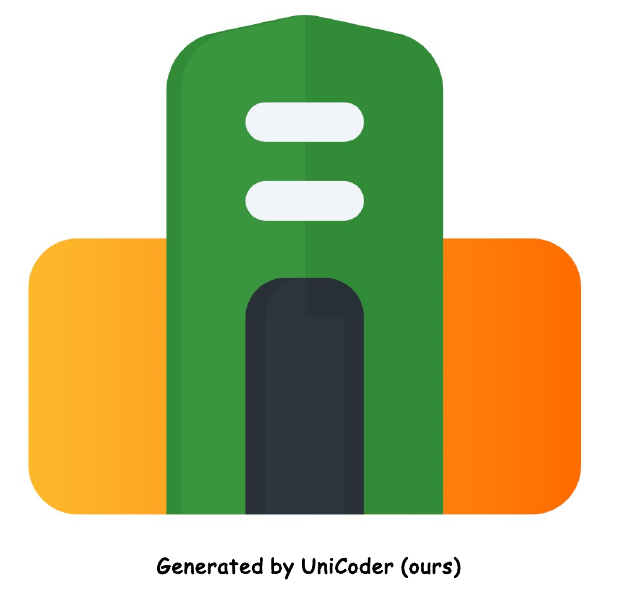}
    \caption{Qualitative comparison between our proposed method and various baselines.}
    \label{fig:case_3_ours}
\end{figure*}

\begin{figure*}[t!]
    \centering
    \includegraphics[width=0.65\textwidth]{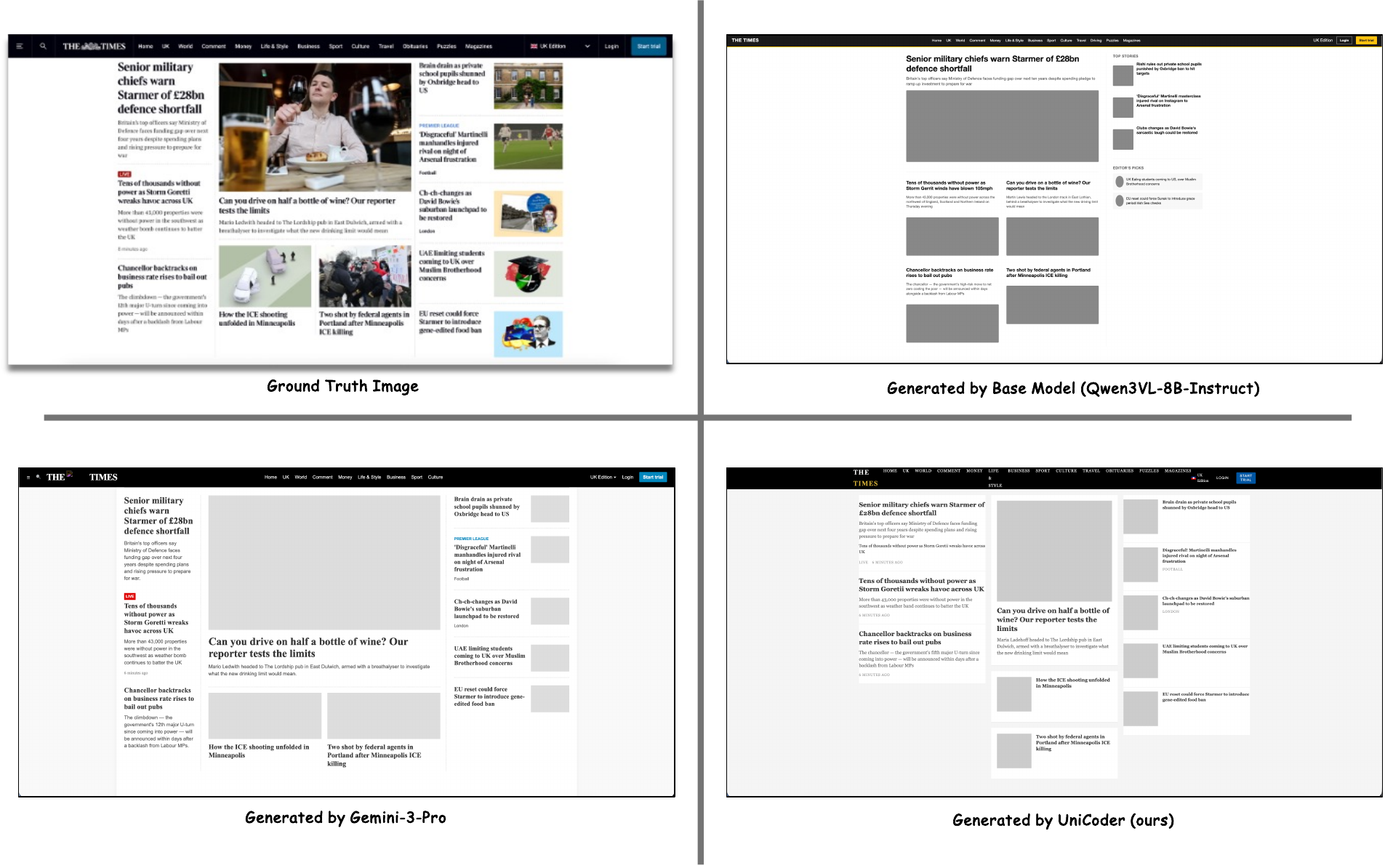}
    \caption{Qualitative comparison between our proposed method and various baselines.}
    \label{fig:case_4_r}
\end{figure*}

These examples (Figure \ref{fig:case_2_r}, \ref{fig:case_2_ours}, \ref{fig:case_3_r}, \ref{fig:case_3_ours}, \ref{fig:case_4_r}) were selected to cover the diverse modalities addressed in this work. They serve to highlight the common failure modes of existing end-to-end MLLMs, such as \textit{color hallucination}, \textit{topological errors}, and \textit{element omission} in complex layouts—and illustrate how our unified, reference-guided approach successfully overcomes these challenges. As shown in the comparisons, our method consistently produces code with significantly higher visual fidelity and structural completeness, more closely matching the precise design intent of the source visuals.

\section{Implementation Detail}

\subsection{Prompt Template}
\begin{figure*}[t!]
    \centering
    \includegraphics[width=0.85\textwidth]{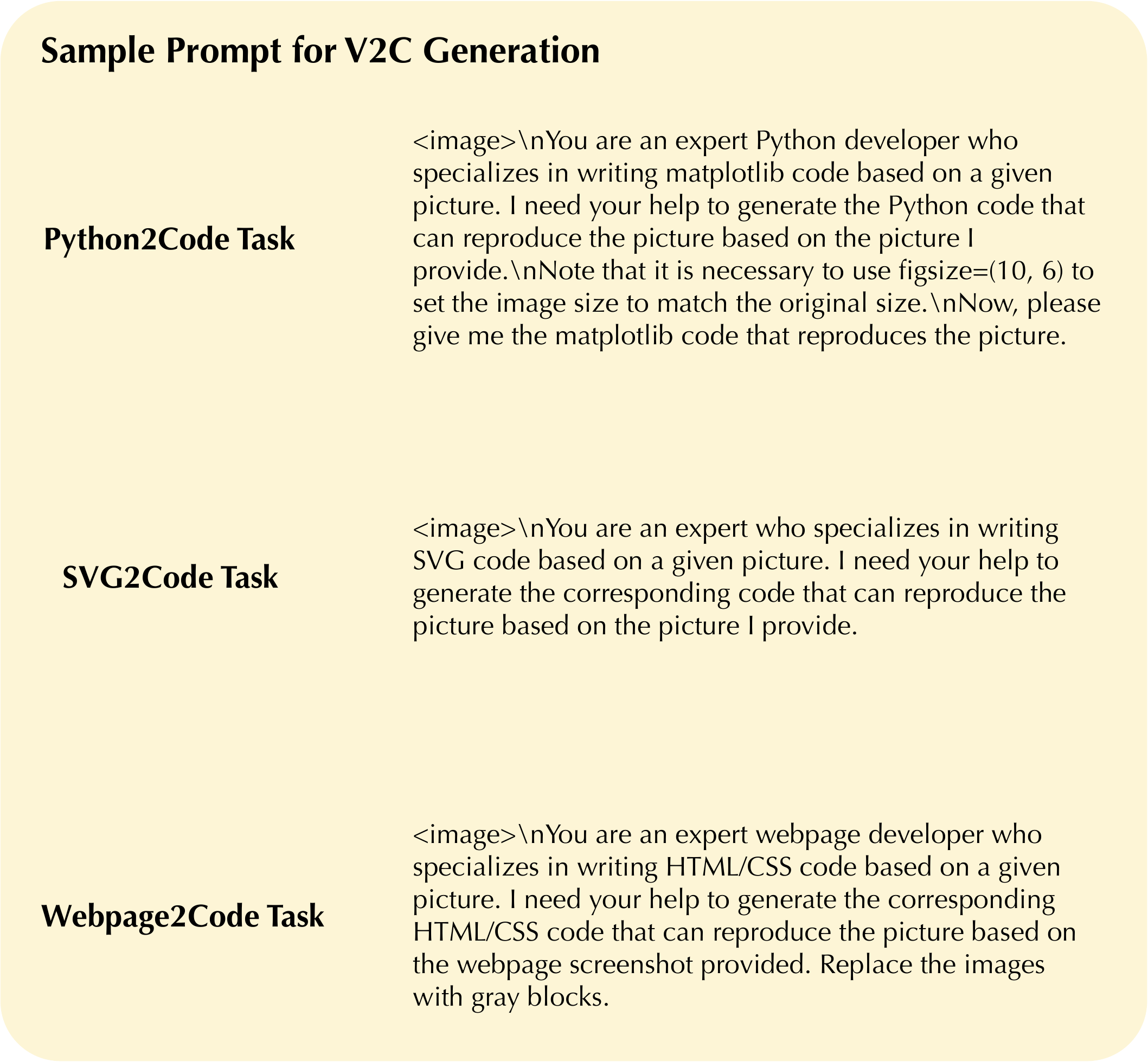}
    \caption{Prompt Templates for Various Code Generation Tasks.}
    \label{fig:prompt1}
\end{figure*}

\begin{figure*}[t!]
    \centering
    \includegraphics[width=0.85\textwidth]{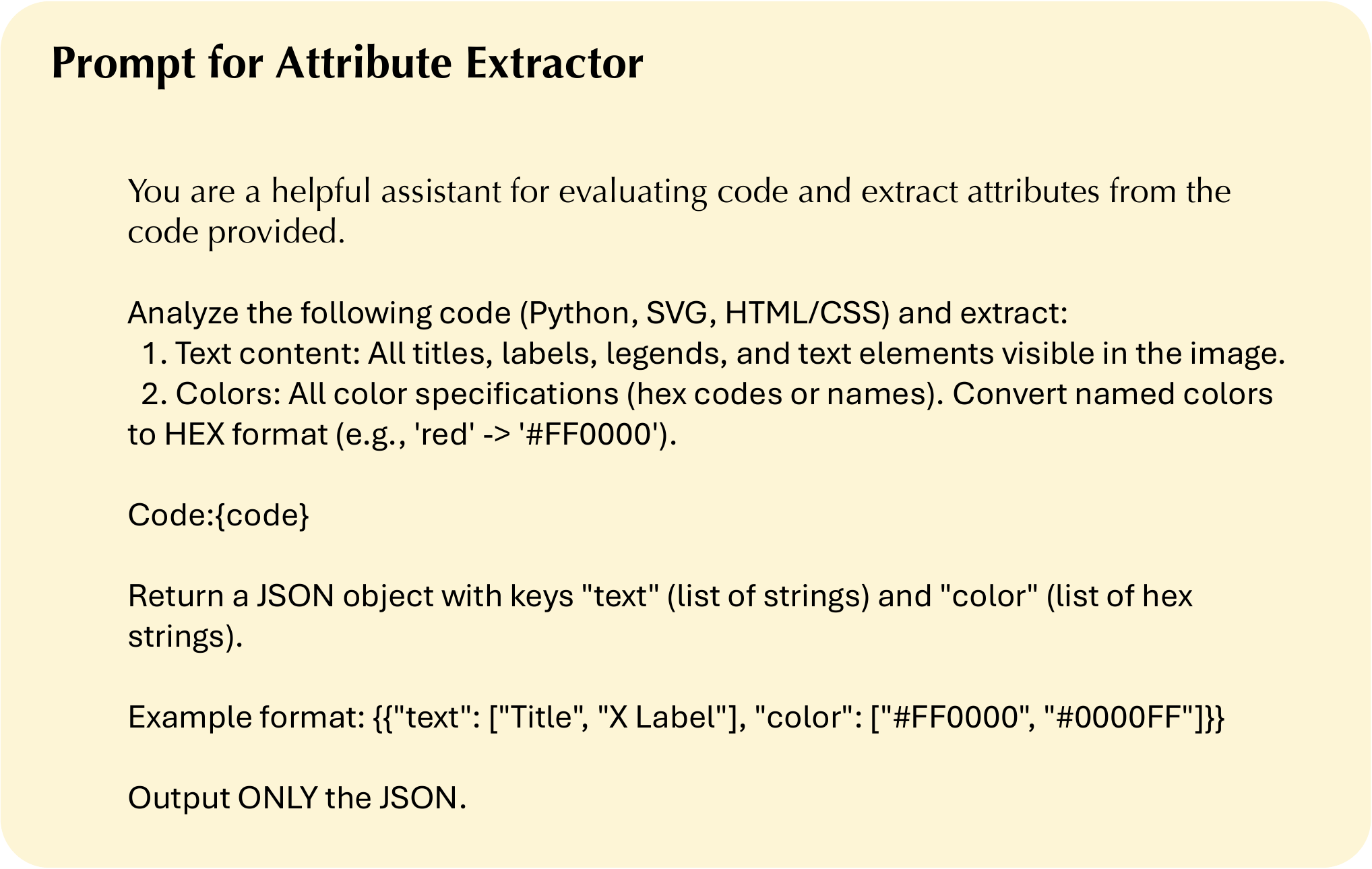}
    \caption{Prompt Templates for Our Attribute Extract Module.}
    \label{fig:prompt2}
\end{figure*}

The prompt templates are shown in Figure \ref{fig:prompt1}, \ref{fig:prompt2}.

\subsection{Training Details}
Our training and experiments were conducted using the EasyR1 codebase on a computational node equipped with 8 NVIDIA A100 GPUs. We adopted a single-stage Reinforcement Learning pipeline. The model, initialized from Qwen3-VL-8B-Instruct, was directly optimized using our proposed algorithm based on Group Relative Policy Optimization (GRPO). The training was performed on a selected subset of 3000 from the MSRL, SVG-Stack, and Web2Code datasets. We configured the training with a global batch size of 128 and a rollout number ($G$) of 16. The model was optimized with a peak learning rate of $4 \times 10^{-6}$ and bfloat16 precision. During inference and rollout generation, we used a temperature of 0.7 with a maximum response length of 4096 tokens.The reward function incorporated our element-level reward and the conventional CLIP score to explicitly align the generated code with visual constraints.


\end{document}